\documentclass[letterpaper, 10 pt, conference]{ieeeconf}  
\pdfoutput=1

\IEEEoverridecommandlockouts                              

\usepackage{amsmath} 
\usepackage{amssymb}  
\usepackage[usenames,dvipsnames]{color}
\usepackage{graphicx}
\usepackage{lipsum}
\usepackage{siunitx}
\usepackage{subfig} 
\usepackage{gensymb} 
\usepackage[short]{optidef} 
\usepackage{bm} 
\usepackage[dvipsnames]{xcolor} 
\usepackage[flushleft]{threeparttable}

\usepackage{tikz}
\usetikzlibrary{positioning}
\usetikzlibrary{shapes.multipart}
\usetikzlibrary{plotmarks}
\usetikzlibrary{arrows.meta}
\usetikzlibrary{shapes,arrows}
\usetikzlibrary{calc}

\DeclareMathOperator*{\argmax}{arg\,max}

\newcommand{\ph}[1]{{\textbf{#1:}}}

\newcommand{\inr}[1]{\in \mathbb{R}^{#1}}

\usepackage{colortbl}
\usepackage{color}

\usepackage{multirow}
\usepackage{graphicx}
\usepackage{booktabs}

\title{\LARGE \bf
Early Recall, Late Precision: Multi-Robot Semantic Object Mapping under Operational Constraints in Perceptually-Degraded Environments}

\author{Xianmei Lei$^{1*}$, Taeyeon Kim$^{2*}$, Nicolas Marchal$^{1*}$, Daniel Pastor$^{1}$, Barry Ridge$^{1}$, Frederik Schöller$^{1}$ \\Edward Terry$^{1}$, Fernando Chavez$^{1}$, Thomas Touma$^{1}$, Kyohei Otsu$^{1}$ and Ali Agha$^{1}$
\thanks{This research was carried out by the Jet Propulsion Laboratory, California Institute of Technology, under a contract with the National Aeronautics and Space Administration.}
\thanks{$^{1}$All authors are with Jet Propulsion Laboratory, California Institute of Technology, United State.}%
\thanks{$^{2}$T. Kim is with the Department of Electrical Engineering, Korea Advanced Institute of Science and Technology, South Korea.}%
\thanks{$^{*}$Equal contribution. Corresponding Authors: xianmei.lei@jpl.nasa.gov, ty.kim@kaist.ac.kr, marchalnico@gmail.com}%
\thanks{\copyright 2022 California Institute of Technology. All rights reserved.}
}

\begin{document}

\maketitle
\thispagestyle{empty}
\pagestyle{empty}

\begin{abstract}
Semantic object mapping in uncertain, perceptually degraded environments during long-range multi-robot autonomous exploration tasks such as search-and-rescue is important and challenging. During such missions, high recall is desirable to avoid missing true target objects and high precision is also critical to avoid wasting valuable operational time on false positives.  Given recent advancements in visual perception algorithms, the former is largely solvable autonomously, but the latter is difficult to address without the supervision of a human operator.  However, operational constraints such as mission time, computational requirements, mesh network bandwidth and so on, can make the operator's task infeasible unless properly managed. We propose the Early Recall, Late Precision (EaRLaP) semantic object mapping pipeline to solve this problem. EaRLaP was used by Team CoSTAR in DARPA Subterranean Challenge, where it successfully detected all the artifacts encountered by the team of robots. We will discuss these results and performance of the EaRLaP on various datasets.

\end{abstract}

\section{INTRODUCTION}
Multi-robot systems with multi-sensor payloads have facilitated a breadth of applications in recent years, ranging from search-and-rescue
operations in which such systems are tasked with autonomously navigating harsh environments to find survivors \cite{morrisRoboticIntrospectionExploration2007}, to potential unmanned exploration of subterranean environments of planets, asteroids and other bodies in our solar system and beyond \cite{aghaRoboticExplorationPlanetary2019}. These developments have been fueled in part by a significant increase in processor efficiency, allowing for advanced neural network architectures and other complex algorithms to be run in real-time aboard robots with significant size, weight and power (SWaP) limitations \cite{reutherSurveyBenchmarkingMachine2019}. In tandem with such advances, the field is also progressing via the diversification of the underlying mobility platforms beyond traditional wheeled systems and towards solutions that are more adapted to certain types of environments. These include relatively fast-paced legged robots traversing difficult and unknown terrain \cite{Bouman2020AutonomousSL}.

Thanks to these developments, a critical point has been reached where it is now possible to run modern, highly capable algorithms for object detection and localization, terrain navigation, hazard avoidance and other purposes on a variety of modestly-sized, agile and low-powered robots with the goal of semantically mapping harsh environments. However, while such ideas are compelling in theory, they can be hampered in practice by a myriad of challenges when the robots are deployed outside of laboratory settings in real-world scenarios. One such challenge is that even state-of-the-art visual detection and localization algorithms can perform sub-optimally when faced with uncontrolled scenarios in which perceptual degradation from motion blur, shifting luminosity, sensor failure, occlusion and other ocular hazards, are the rule rather than the exception. Another significant challenge is posed by communications constraints. The low-bandwidth wireless mesh networks that are typically employed by such multi-robot systems demand that limits be placed on the size and frequency of visual observation reports sent by robots to an operational base station.

\begin{figure}[!t]
  \centering
  \includegraphics[width=1\linewidth]{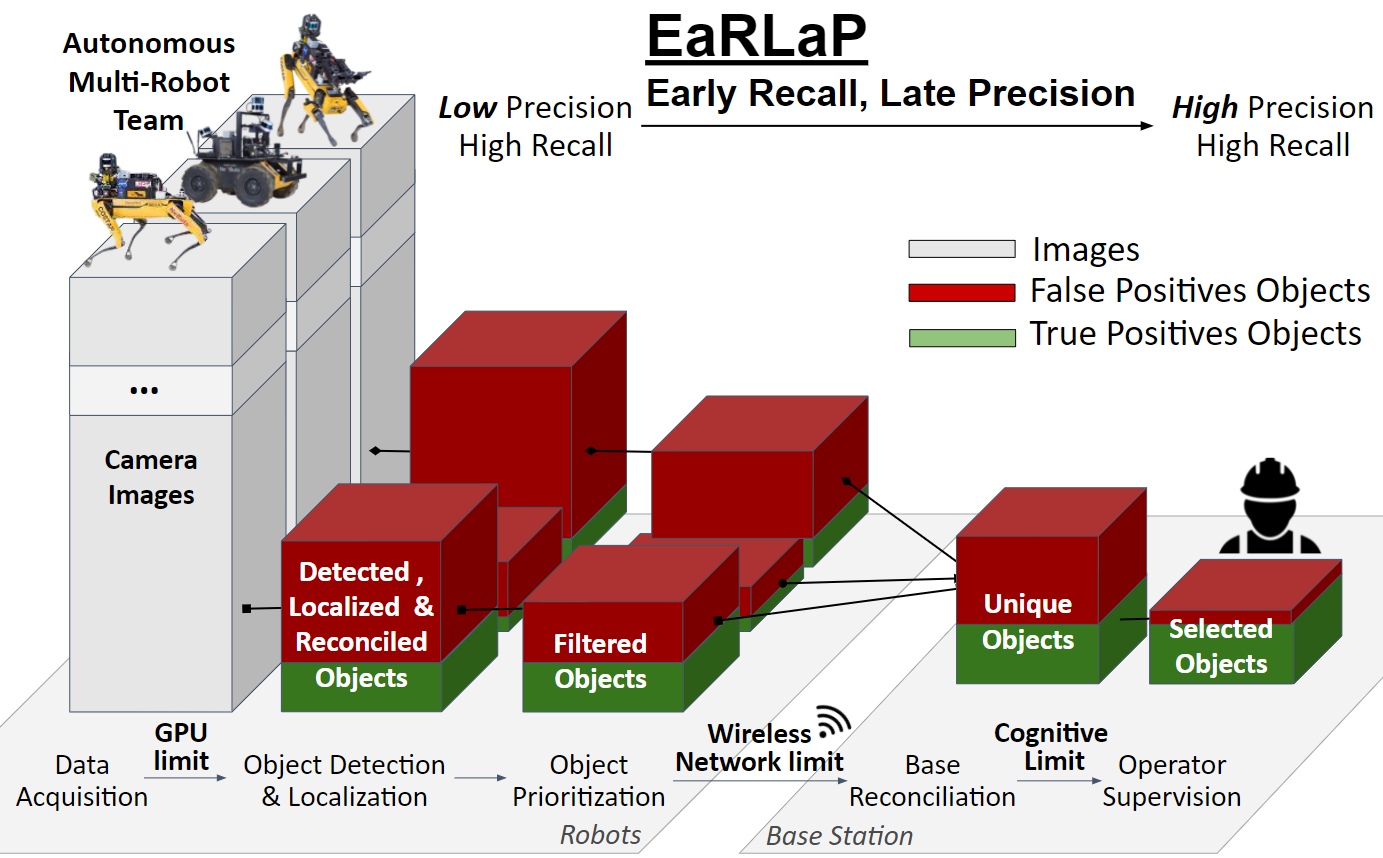}
  \caption{\footnotesize Early Recall, Late Precision (EaRLaP) semantic object mapping pipeline. Objects detected with high recall in robot-gathered images are localized, then false positives are pruned in multiple stages to respect network constraints before transmission to a base station. After reconciliation of unique objects, an operator reviews detections to achieve high precision.}
  \label{fig:arop}
  \vspace{-0.4cm}
\end{figure}

Depending on the particular application, perceptual degradation may not always be such a concern, however in tasks involving object detection and localization in which relatively high rates of precision and recall are required, severe performance gaps can emerge. In the search-and-rescue example, it is desirable to both always detect real survivors when they are encountered (high recall) such that all survivors are rescued, and to only report detections that really are survivors (high precision) to avoid hampering rescue efforts. While it is possible to trade off precision for recall, or vice versa, \cite{buckland1994relationship} it can be exceptionally difficult to achieve high rates of both in perceptually degraded circumstances. One possible way to bridge this performance gap is to leverage a visual detection algorithm to achieve high recall and then introduce the supervision of a human operator to make up the precision deficit by identifying true survivors from the detections reported by the algorithms. The downside to this approach is that when the recall is high but the precision is low, the ratio of false positives to true positives that are initially reported might be extremely high and this has the potential to overwhelm both the communications network and the cognitive capacity of the operator in time-critical scenarios.

We present the \emph{Early Recall, Late Precision} (EaRLaP) semantic object mapping pipeline that attempts to optimize human-in-the-loop multi-robot object detection and localization performance under various operational constraints. EaRLaP, illustrated in Fig. \ref{fig:arop}, initially maximizes recall to ensure that no objects are missed and then siphons the large number of detections through a series of filtering stages that gradually prune false positives and compresses information to respect various constraints and achieve high precision. EaRLaP is resilient against uncertainties encountered in the real world (sensor failures, unknown environments, harsh conditions, etc.) and has been tested in extremely challenging situations by the \emph{NASA Jet Propulsion Laboratory (JPL)}. EaRLaP has been integrated as a critical part of the \emph{Networked Belief-aware Perceptual Autonomy (NeBula)} package \cite{agha2021nebula} and was used by JPL team CoSTAR during the final event of the DARPA SubT challenge.

The DARPA Subterranean Challenge (SubT) was a 3-year competition organized by DARPA, that sought novel approaches to rapidly map, navigate, and search underground environments \cite{rouvcek2019darpa}. The goal was to find as many artifacts as possible within a set time limit, and only a very limited number of submissions could be made to the scoring system for evaluation. The artifacts consisted of a fixed set of objects associated with search-and-rescue scenarios, such as backpacks, ropes and helmets, as well as other entities that might not fit the usual object definition, such as RF signals, and CO2 gas. A critical challenge laid in managing the quantity and quality of the information presented to the operator.
\ph{Contribution} 
EaRLaP is an extension of our previous publication \cite{terryObjectGasSource2020} with several new key contributions. First, we provide the mathematical formulation of both our problem and solution. The EaRLaP pipeline is thus anchored in a mathematical framework which justifies its structure, while the pipeline in \cite{terryObjectGasSource2020} was developed through intuition. Additionally, EaRLaP introduces new steps in the pipeline to solve the precision issues discussed in \cite{terryObjectGasSource2020} and updates the existing stages with an enhanced hardware setup and new algorithms to address the reliability issues discussed in \cite{terryObjectGasSource2020}. Finally, this work provides a quantitative analysis of our performance during the final event of the DARPA SubT Challenge, where we have demonstrated some of the strongest semantic object mapping performance.

In the remainder, we discuss related work in Sec. \ref{sec:related_work}, define the problem formulation in Sec. \ref{sec:problem_description}, detail the methodology of the EaRLaP approach in Sec. \ref{sec:methodology}, and describe results from real in-field experiments and SubT final event competition runs in Sec. \ref{sec:exp_res}, before concluding in Sec. \ref{sec:conclusions}.

\section{RELATED WORK}
\label{sec:related_work}

\ph{Perceptually-Resilient CNNs}
In perceptually degraded environments, convolutional neural networks (CNNs) suffer from reduced performance due to significant differences between training data and low quality test images. Combining thermal and visual images can enable a system to be robust to low light conditions and obscurant-filled settings (e.g. dust filled) \cite{tsiourva2020Multi}. To overcome motion blur, it has been suggested to add deblurring models \cite{xie2021Plug} into the detection pipeline. A deblurring model can be pre-trained on data blurred with directional kernels and included without retraining the detection model. To robustify a detector that operates in a setting which differs from its training data, \cite{khodabandeh2019robust} tried domain adaptation and \cite{marchal2020learning} used normalizing flows. Finally, to completely overcome challenges due to poor image quality, point clouds detectors can be trained using adversarial training\cite{debortoli2021adversarial}. While all of the approaches above are interesting, due to limited computational power on our robots we had to use a simple but efficient single stage network, such as YOLO \cite{yolov5}.

\ph{Object Localization}
To localize an object, a robot must first estimate its own position, which can be done with SLAM if no prior map of the environment is available \cite{lamp, locus}. Robots must then estimate the relative position of the object, which is most commonly done by using a sensor that provides both color and depth information \cite{terryObjectGasSource2020}. Stereo cameras are a popular choice, but projecting LiDAR data onto images to obtain a depth map is much more accurate at long ranges \cite{agrawal2019ground}.  An alternative to using depth sensors is to either create a depth map with monocular depth neural networks \cite{godard2019digging} or to directly predict an object's range with a specialized network \cite{godard2019digging}. Given an observation direction and distance, one can leverage existing libraries to compute the position of the object \cite{dellaertFactorGraphsGTSAM2012a}. 

\ph{DARPA SubT Challenge}
Different solutions for semantic object detection and localization in degraded conditions have been proposed for the DARPA SubT Challenge, where both object detection and communication bandwidth limits are challenging. Team MARBLE \cite{ohradzanskyMultiagentAutonomyAdvancements2021} relied on a YOLOv3 architecture \cite{Redmon2018YOLOv3AI} for object detection. They handled limited bandwidth by using deployable relay nodes to extend range together with each robot being able to explore independently and report its findings once within range of the communications network. Team CERBERUS \cite{tranzattoCERBERUSAutonomousLegged2021} used a combination of YOLOv3 \cite{Redmon2018YOLOv3AI} and manual selection for object detection. Communication was handled by a mesh network consisting of deployable WiFi nodes. The mesh network was extended by maneuvering a remote controlled rover equipped with a high gain WiFi antenna connected to the base station by a 300m fiber optical cable. Team CSIRO \cite{hudsonHeterogeneousGroundAir2021} used the DeNet \cite{tychsen2017denet} object detector and implemented a navigation algorithm to give their robots the ability to return within communication range to report back observations to the base. Team EXPLORER \cite{schererResilientModularSubterranean2021} used simulation data in addition to real-world data to train their detectors.


\section{PROBLEM DESCRIPTION}
\label{sec:problem_description}
In this section, we describe a mathematical formulation of the semantic object mapping problem and illustrate how to define the problem as a theoretical optimization problem.  We then describe why such an optimization is infeasible in closed-form, and illustrate the approximated solution constructed by breaking the problem into a series of constrained sub-problems solved by the proposed EaRLaP approach.

\subsection{Semantic Object Mapping Problem (SOMP)}
\label{subsec:som}
The goal of the semantic object mapping problem is to detect and localize a
set of objects using a mobile robot autonomously navigating an unknown environment. The problem is defined using a list of $G$ ground truth object tuples $\mathbf{g}=[\{\mathbf{p}_i^g,l_i^g\}_{i=1}^G]$,
named the ground truth set. Given an element of the list $g_i =
\{\mathbf{p}_i^g,l_j^g\}$, $\mathbf{p}_i^g\inr{3}$ is the 3D position of the
ground truth object $i$ and $l_i\in[0,\dots,L]$ is its semantic label, selected
from a set of $N$ possible labels. The desired output of the proposed semantic
object mapping algorithm is a similar list of $S$ tuples
$\mathbf{s}=[\{\mathbf{p}_i,l_i\}_{i=1}^S]$ called the submission set. Each
submitted position is compared with the closest ground truth position of the
same label to compute the object localization error:
\begin{align}
    e_i(\mathbf{g},s_i)=\min_{\mathbf{p}^g_j\in \mathbf{g}}|\mathbf{p}_i-\mathbf{p}^g_j|,~~~s.t.,~~l_i= l_j.
\end{align}

If an object is correctly classified and its localization error is less than
the acceptable limit $e_{\max}$, the object is considered detected and a positive
reward  $\mathsf{R}_i(e_i) = U(e_i\leq e_{\max})$ is received, where $U(\cdot)$ returns 1 when input is true and 0 otherwise.
The total reward for the whole submission set is $   \mathsf{R}(\mathbf{g},\mathbf{s}) = \sum_{i=1}^S   \mathsf{R}_i(\mathbf{g},s_i)$.

The general semantic object mapping problem considered is defined as follows - given $S_{max}$ as the maximum allowed number of submissions, find the best submission set that maximizes the detection reward:
\begin{align}
    \mathbf{s}^* = \argmax_{\mathbf{s}} \mathsf{R}(\mathbf{g},\mathbf{s}),~~~s.t.,~ |\mathbf{s}|\leq S_{\max}.
    \label{eq:reward}
\end{align}

\subsection{Multi-Robot SOMP}
Assuming that there are $R$ robots connected to a single base station via a mesh network.
Each robot receives an RGB-D measurement $\mathbf{I} = [\mathbf{I}_{\text{RGB}}, \mathbf{I}_{\text{d}}]$, where $\mathbf{I}_{\text{RGB}}$ is an RGB image and $\mathbf{I}_d$ is the depth image obtained by a depth sensor (e.g., a LiDAR or stereo camera). Each camera takes an image at rate $f$ for $T$ seconds, creating $2fRT$
images. This defines the vector of all images $\mathcal{I}$. Note that it is a vector and
not a set as the ordering of the images can be exploited by the object detection
algorithm. 
We define the function $f$ mapping a vector of images $\mathcal{I}$ to the submission set $\mathbf{s}$ parametrized with $\theta \inr{N_{\theta}}$
\begin{align}
\mathbf{s}&=\{p_i,l_i\}_{i=0}^S 
            =f(\mathcal{I};\theta).
\label{eq:f0}
\end{align}


The parameters $\theta$ could be learned from a dataset of the form $[\mathcal{I},\mathbf{g}]$ containing image vectors and associated ground truth positions and labels. An agent learning the solution would need to perform the following optimization:
 \begin{align}
   \theta^* = \argmax_{\theta} \mathsf{R}(\mathbf{g},\mathbf{s}) 
                   \quad \textrm{s.t.} \:\:\: \mathbf{s} = f(\mathcal{I};\theta), \:\:\: |\mathbf{s}| \leq  S_{\max}.
\label{opt:theta_0}
 \end{align}



First, the reward function in \eqref{opt:theta_0} is discontinuous and the input space is very high dimensional, which makes it difficult to optimize. Additionally, in practice, and within the confines of the DARPA SubT challenge in particular, the problem is subject to a number of additional constraints and
nuances that make the direct optimization of $\theta$ infeasible.
Thus, in the following Sec. \ref{sec:sub_problems}, we describe these
additional constraints as well as how the general problem can be divided into tractable sub-problems that can be solved by our proposed EaRLaP pipeline
approach.

\section{Constrained Sub-Problem Decomposition}
\label{sec:sub_problems}

In the robot-and-base-station scenario, robots must perceive the environment and communicate their findings with a base station and we face trade-offs in deciding how to distribute the computational aspects of this perceptual pipeline between them. Depending on the computational capabilities and network bandwidth, it is necessary to perform some operations on the robot, at the base station, or both. In addition, we exploited the supervision of a human operator to create a final high-confidence submission set $\mathbf{s}$, but the operator can only verify a limited number of detections within the allotted mission time.
With these constraints, we decompose the function $f$ previously defined in \eqref{eq:f0} into three sub-functions - $f_r$ for the robot, $f_b$ for the base station and $f_{op}$ for the operator:

\begin{align}
\begin{split}
  f(\mathcal{I};\theta) &= f_{op} \circ f_b\circ f_{r}(\mathcal{I}),~~~\theta=(\theta_r, \theta_b, \theta_{op}), \\
  \mathbf{s} &= f(\mathcal{I}, \theta) = f_{op}\left( f_b\left( f_r\left(\mathcal{I};\theta_r\right);\theta_b\right);\theta_{op}\right).
  \end{split}
  \label{eq:partly_separated}
\end{align}


In this decomposition,
the data $\mathcal{I}$ is first processed on the robot using $f_r(\mathcal{I}, \theta_r) = \mathbf{s}_r = \mathcal{I}_r$ such that the amount of data $|\mathcal{I}_r|$ being transmitted from the robot to the base station is within the communication bandwidth limit $S_b$, i.e. $|I_r| < S_b$. Finding the optimal $\theta_r$ thus follows the same manner of SOMP problem \eqref{opt:theta_0} using the mathematical formulation:
\begin{align}
  \theta_r^* \:=\: &\argmax_{\theta_r} \mathsf{R}(\mathbf{g},\mathbf{s}_r) \nonumber \\
                 \:\textrm{s.t.} \quad &\mathbf{s}_r = f_r(\mathcal{I};\theta_r), \quad |\mathbf{s}_r|  \leq  S_{b}.
\label{opt:theta_r}
\end{align}

\begin{figure*}[!ht]
\centering
    \resizebox{\textwidth}{!}{%
\begin{tikzpicture}[
robotnode/.style={rounded corners=0.2cm, draw=black!60, fill=green!5, , minimum size=15mm, align=center},
basenode/.style={rounded corners=0.2cm, draw=black!60, fill=gray!5, , minimum size=15mm, align=center},
humannode/.style={rounded corners=0.2cm, draw=black!60, fill=red!5, , minimum size=15mm, align=center},
topnode/.style={rounded corners=0.2cm, draw=black!60, fill=red!0, , minimum size=15mm, align=center},
squarednode/.style={circle, draw=black!60, fill=red!0, , minimum size=10mm, align=center},
]
\node[squarednode,draw]      (SensorData)         {Sensor \\ Data: $\mathcal{I}$};
\node[robotnode,draw]        (cnn)       [right=of SensorData, label=below:$f_1$] {Object \\  Detector};
\node[robotnode,draw]      (ColorFilter)       [right=of cnn, label=below:$f_2$] {Color \\ Filter};
\node[robotnode,draw]        (robotrecon)       [right=of ColorFilter, label=below:$f_3$] {Robot \\ Reconciler};
\node[robotnode,draw]        (comms)       [right=of robotrecon, label=below:$f_4$] {Communication \\ Filter};
\node[basenode,draw]        (baseRecon)       [right=of comms, label=below:$f_5$] {Base \\ Reconciler};
\node[basenode,draw]        (confidence)       [right=of baseRecon, label=below:$f_6$] {Scorability \\ Ranker};
\node[humannode,draw]        (humanFilter)       [right=of confidence, label=below:$f_7$] {Human \\ Filter};
\node[squarednode,draw]        (solution)       [right=of humanFilter] {Solution: \\$\mathbf{s}$};

\begin{scope}[yshift=3cm]
\node[robotnode,draw,text depth=0.35ex]        (robot)       [above=of robotrecon, label=north :$f_r$] {Robot};
\node[basenode,draw,text depth=0.35ex]        (baseStation)     at(15.2,-0.5)  [ label=north west:$f_{b}$] {Base \\ Station};
\node[humannode,draw,text depth=0.35ex]        (human)       [above=of humanFilter, label=above:$f_{op}$] {Human};
\end{scope}

\node[topnode,draw]        (semanticmapping)       [above=of baseStation, label=north :$f$] {Semantic \\ Mapping};

\draw[-stealth] (SensorData) -- (cnn);
\draw[-stealth] (cnn) -- (ColorFilter);
\draw[-stealth] (ColorFilter) -- (robotrecon);
\draw[] [dashed](6.2,-2) -- (6.2,0.9);
  \node[text width=3cm] at (7.8,-1.7) 
    {\small Detection Output};
\draw[-stealth] (robotrecon) -- (comms);
\draw[-stealth] (comms) -- (baseRecon);
\draw[-stealth] (baseRecon) -- (confidence);
\draw[-stealth] (confidence) -- (humanFilter);
\draw[-stealth] (humanFilter) -- (solution);

\draw[-stealth] (robot) -- (cnn);
\draw[-stealth] (robot) -- (ColorFilter);
\draw[-stealth] (robot) -- (robotrecon);
\draw[-stealth] (robot) -- (comms);

\draw[-stealth] (robot) -- node[anchor=south]{$I_r$}(baseStation);

\draw[-stealth] (baseStation) -- (baseRecon);
\draw[-stealth] (baseStation) -- (confidence);

\draw[-stealth] (human) -- (humanFilter);
\draw[-stealth] (baseStation) -- node[anchor=south]{$I_b$}(human);
\draw[] [dashed](12.5,-2) -- (12.5,4.5);
  \node[text width=3cm] at (14.1,-1.7) 
    {\small Robot Output};

\draw[-stealth] (semanticmapping) -- (robot);
\draw[-stealth] (semanticmapping) -- (baseStation);
\draw[-stealth] (semanticmapping) -- (human);
\draw[] [dashed](18.1,-2) -- (18.1,4.5);
  \node[text width=3cm] at (19.7,-1.7) 
    {\small Base Output};

\end{tikzpicture}
}
  \caption{\footnotesize Hierarchical decomposition of the semantic object mapping problem. $f$ is divided into three sub-functions and these sub functions are further divided to reach a composition of seven sub-functions, which is the EaRLaP semantic object mapping pipeline.}
  \label{fig:prob_decomposition}
\end{figure*}
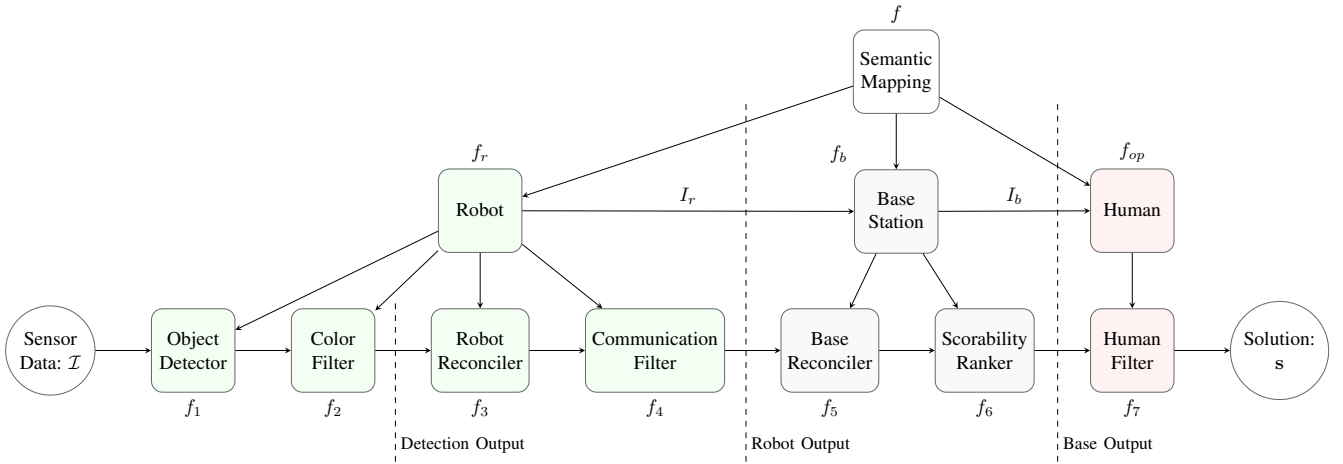

The amount of transmitted data $|\mathcal{I}_r|$ on the base station is assumed to be greater than the maximum number of submissions $S_{max}$. We therefore further process $\mathcal{I}_r$ on the base station  with $f_b(\mathcal{I}_r, \theta_b) = \mathcal{I}_{b}$ to condense the data into $\mathcal{I}_b$, such that $|\mathcal{I}_b|$ is small enough for a human operator to review within time limit. The maximum size of $\mathcal{I}_b$ is governed by the maximum human cognitive load $S_{op}$ or maximum trained human operator cognitive load which is also assumed to be greater than $S_{max}$. 
Once again, finding the $\theta_b$ is an optimization problem of the same form as \eqref{opt:theta_0}.

Finally, the operator is responsible to manually process $\mathcal{I}_b$ to create the set of submissions $s = f_{op}(\mathcal{I}_b, \theta_{op})$, and ensure that $|\mathbf{s}| < S_{max}$. This follows the formulation \eqref{opt:theta_0}, but $f_{op}$ does not follow a mathematical formulation since $\theta_{op}$ represents the operator's human brain process capability. Note that this is still an optimization problem, and the operator undergoes training to optimize the detection performance.

With this high-level decomposition of $f$ in place, we can further decompose the sub-functions $f_{r}$, $f_b$ and $f_{op}$ hierarchically into a set of $N$ tractable lower-level sub-functions $f_{i}(\mathcal{D}_{i-1};\theta_{i})$, each of which permit similar optimization formulations as \eqref{opt:theta_r}, which are then combined to give a solution to the original problem:
\begin{equation}
    f(\mathcal{I};\theta) = f_N \circ\cdots\circ f_{i} \circ\cdots\circ f_1(\mathcal{D}_0;\theta_1), \\
    \label{eq:pipeline}
\end{equation}
where $\theta = (\theta_1,\ldots,\theta_N)$ provides parameterizations for each sub-function, $\mathcal{D}_{0} = \mathcal{I}$, and $\mathcal{D}_{i} = f_{i}(\mathcal{D}_{i-1};\theta_{i})~\forall i=1,\ldots,N$ are data outputs that may include images, point clouds etc. and/or higher-level semantic information such as bounding boxes and labels, depending on the sub-function in question.

\section{EaRLaP Semantic Object Mapping Pipeline}
\label{sec:methodology}

The implementation of \eqref{eq:pipeline} that was designed for the EaRLaP pipeline and ultimately deployed in the final events of the DARPA SubT Challenge consisted of $N=7$ different sub-functions as illustrated in Fig. \ref{fig:prob_decomposition} and detailed below.


\ph{$f_1$ - Object Detector}
The sensor payloads on the robots contain multiple cameras to cover the entire field of view (FOV), however this entails that multiple high-rate image streams must be handled by CNN object detectors at a low rate on robot processors with limited computational power. To ensure computational power is used efficiently, an image selection filter samples a subset of images with higher quality.
To obtain high-quality detections, in particular for small objects in low lighting environments, high-resolution images \cite{data_aug} must be used. Lightweight models designed for real-time application and high-resolution images, such as the family of YOLO models \cite{yolov5}, are suitable for $f_1$.  

Given an image $\mathbf{I}$, the object detector outputs $N_d$ detections with associated labels, confidences and bounding boxes $\{l, c, \mathbf{B}\}_{i=1}^{N_d} = g_{\textrm{yolo}}(\mathbf{I})$. For each label $l$ the training data is used to set a threshold $t_l$ that maximizes the F-score. Only the detections with label $l$ and confidence higher than $t_l$ are passed on to the next sub-function. 

\ph{$f_2$ - Color Filter} 
After the CNN step, to remove false-positive detections, a function that extracts handcrafted features can be applied if an object types contain outstanding features, such as color and shape. In the SubT case, the object types and their respective coloring are known, object detection results with unexpected color patterns can be eliminated.
To identify the color of an object, the segment of the RGB image $\mathbf{I}$ inside the detection bounding box $\mathbf{B}$, that is $\mathbf{I}_\mathbf{B}$, is transformed to HSV color space and a mask $\mathbf{M}_l$ is applied which is tailored to the color of each object $l$. Note that in practice $\mathbf{M}_l$ must accept a wide spectrum of color to account for the different conditions in which we might encounter an object. The masked bounding box is thresholded and the remaining pixels are counted to obtain an object color score percentage $p$.  Each object $l$ has a multiplier $\beta_l > 1$ such that the color score defined by $g_{\textrm{col}}(l, \mathbf{I}_\mathbf{B}) = \min(\beta_l \cdot p, 1) \in [0, 1]$ returns 1 if even only part of $\mathbf{I}_\mathbf{B}$ is the correct color. To filter detections, the color score is multiplied with the yolo confidence, and the same YOLO threshold $t_l$ is used to decide if a detection moves on to the next step.

\ph{$f_3$ - Robot Reconciler}
The global position of a given robot relative to a calibration gate at its starting point is estimated by other system modules \cite{lamp, locus}, however the detected objects still need to be reconciled and localized relative to the robot.  In order to initially eliminate noisy detections, when a detection with label $l$ is received, it is removed if the robot has not moved at least $d_{min}$ meters or rotated $\alpha_{min}$ degrees since the last observation of another object $l$. The remaining detections must then be reconciled with previously mapped object candidates.

Given an observation, those candidates that are not compatible with the bearing of the observation are not considered (a detection from the front camera cannot come from a candidate located behind the robot, for example). Candidates further than a set maximum distance $d_{\max}^d$ from the robot are also filtered, since detection performance drops significantly beyond that point. If there are multiple candidates after this initial stage, matching the detection with a candidate that is close to the robot and close to the detection direction is favored. After matching the detection to a candidate, the candidate's position and metadata is updated to keep some information about the detection (i.e. its detection confidence and color confidence). If there are no remaining candidates after this process, a new object is created in the map, which will become a new candidate for reconciliation of subsequent object detections.

The primary means used for robot-relative object position localization are bearing estimates (implemented using the GTSAM library \cite{dellaertFactorGraphsGTSAM2012a}), but the estimates can have high variance, particularly with respect to object range. To refine the range, one could exploit depth estimates from RGB-D sensors,  however this can quickly lose accuracy or fail completely at large distances depending on the lighting conditions.  Instead, we make use of an alternative approach that transforms LiDAR-derived 3D point clouds into the camera frame in which the object has been detected, re-projects the 3D LiDAR points to 2D image pixels, determines which of
those points/pixels lie within the detected object bounding box $\mathbf{B}$, and calculates their mean distance in the camera frame.
This approach relies on good LiDAR-camera extrinsic calibration, but provides more accurate results at distances outside of the effective range of RGB-D sensors.

Given a calibrated camera and an object detection with associated bounding box $\mathbf{B}$ and estimated range $d$, we can project $\mathbf{B}$ from 2D to 3D to obtain the estimated object size. If an object of class $l$ has well-defined dimensions (e.g. a backpack of size 30x30x30cm), a size score $g_\textrm{size}(l, \mathbf{B}) \in [0, 1]$ can be created that measures how well $\mathbf{B}$ matches the object size.  Similarly to the color score, this must account for a wide range of sizes observed in practice, mainly due to noisy measurements of $d$ and inaccurate $\mathbf{B}$.

\ph{$f_4$ - Communications Filter}
To limit the quantity of reports sent from the robots to the base station, we only transmit reports in which multiple observations have been reconciled. Additionally, reports for which the median report confidence of all reconciled observations is high are favored.

\ph{$f_5$ - Base Reconciler}
In practice, the reconciliation of $f_3$ does not always work perfectly and a robot might send multiple reports of the same object to the base station. Additionally, when multiple robots communicate to the base station, they could report the same objects if they explore the same areas. We can merge these detections into a unique object. A report is grouped to an existing report cluster if it is spatially closer than $d^b_{\textrm{min}}$ to an existing report in that cluster of the same object class. Such distance-based reconciliation assumes that objects of interest are sparsely distributed in the environment. However, false-positive detections can be located close to each other and can be merged into one object during this step. Additionally, when exploring multi-floor buildings, objects on different floors can be reconciled together if $d^b_{\textrm{min}}$ is not chosen wisely.

\ph{$f_6$ - Scorability Ranker}
In order to mitigate against a time-pressured operator missing true positives while evaluating the reconciled reports on the base station, it is prudent to attempt to rank the reports in terms of their scoring potential.
A detection $D = \{l, c, \mathbf{B}\} \in g_\textrm{yolo}(\mathbf{I})$ produced by running the detector on a given image $\mathbf{I}$ is assigned a \emph{scorability confidence}
\begin{equation}
g_D(D) = c \cdot g_{\textrm{col}}(l, \mathbf{I}_\mathbf{B}) \cdot g_{\textrm{size}}(l, \mathbf{B}).
\end{equation}
The \emph{scorability of a report} $R = \{D_i\}_{i=1}^{n_d}$ is then computed as 
\begin{equation}
g_R(R) = \alpha(n_d) \cdot \frac{1}{n_d} \sum_{i=1}^{n_d} g_D(D_i),
\end{equation}
where $\alpha(n_d)$ is a function of the number of detections $n_d$ per reconciled report, used to penalize reports with a low number of observations, and the reports are ranked with respect to $g_R$ in descending order.

\ph{$f_7$ - Human Filter}
A graphical user interface (GUI) was developed that allows the human operator to control robotic exploration behaviour and perform certain actions.
Once the detections from $f_1$ have been reconciled as objects and then ordered using $f_6$, the operator can review them and manually decide whether or not to include them in the final submission set $\mathbf{s}$ using the GUI.  They may also manually make adjustments as necessary in cases of poor reconciliation, etc.


\section{Experimental Results}
\label{sec:exp_res}

This section will showcase the results obtained by Team CoSTAR who implemented EaRLaP to detect and locate objects during the Final Event of the DARPA SubT Challenge. 

Although Team CoSTAR detected objects using a wide range of sensors, the EaRLaP methodology presented in Sec. \ref{sec:methodology} was motivated primarily by visual detection. This section will discuss the results applied to the following RGB-detectable objects from the competition: backpack, drill, fire extinguisher, helmet, cube, rope and survivor.   

Team CoSTAR deployed Husky and Spot mobility platforms: commercially available wheeled vehicle and quadrupeds, respectively. A customized payload that comprised of sensors and computing processors was developed to facilitate high-level autonomy \cite{Bouman2020AutonomousSL}. The payload includes five Intel Realsense D455 cameras, arranged on the robots to provide an near 360\degree~FOV. The Realsense cameras were connected directly to an Nvidia Jetson AGX Xavier, used to run custom camera drivers, the YOLO object detector ($f_1$) and the color filtering ($f_2$). The data was later transmitted to an Intel NUC computer to perform the reconciliation ($f_3$) and select the objects to send to the base via the wireless communication ($f_4$). The robots carried and deployed communication nodes to establish a communication mesh to connect to the base station. The base is a custom-made computer with multiple GPUs and is connected to a monitor displaying the GUI where the operator monitored the mission and interacted with the robots.   

Different methods and parameters were selected for each sub-function for the final event of the DARPA SubT Challenge. For the image selector, we used the highest variance of the Laplacian \cite{bansal2016blur} to compute and select the least blurry images in the 25Hz  image streaming queue. Team CoSTAR's robots' top speed is 1m/s and we aimed for detections at least once every 25cm. This required the model to run at least 5hz per camera. To achieve this, in $f_1$, we used a YOLOv5m6 \cite{yolov5} model with input size of 1280x768 was purned using channel pruning, decreasing the model size by $22.5\%$. The \emph{SiLU} activations were replaced with \emph{ReLU}, allowing quantization (MinMax strategy) for $\text{INT}8$ inference. Finally, the model was hardware optimized with TensorRT. For communications filter $f_4$, regardless of the total number of observations reconciled in a report, a maximum of 4 compressed images (brightest, least blurry, highest report confidence, and closest) were kept for inspection by the human operator in order to further minimize the amount of transmitted data.

\subsection{Evaluation Criteria}
\label{subsec:eval_criteria}
The normative interpretation of precision/recall metrics in a computer vision problem involves, for example, running an object detector on a given image dataset $\mathcal{I} = \{\mathbf{I}_i\}_{i=1}^l$ to produce a set of detections $\mathcal{D}_\mathcal{I} = \{D_i\}_{i=1}^m$ and comparing the $D_i$ to a ground truth set $\mathcal{G}_\mathcal{I} = \{\mathbf{G}_i\}_{i=1}^n$ to produce true positive ($\mathit{TP}$), false positive ($\mathit{FP}$) and false negative ($\mathit{FN}$) counts such that $\mathit{TP} + \mathit{FP} = m$ and $\mathit{TP} + \mathit{FN} = n$.
In this case, $n$ is counted across all images in $\mathcal{I}$.
In our scenario, we are not only interested in such \emph{image-based} metrics, but also in \emph{object-based} metrics.
To help explain this, we introduce the notion of an environment $\mathcal{E}$ in which the physical ground truth objects $\mathcal{G}_\mathcal{E} = \{\mathbf{G}_i\}_{i=1}^k$ are positioned and within which all images $I_i \in \mathcal{I}$ are gathered such that $k \le n$.  By reconciling the $D_i \in \mathcal{D}_\mathcal{I}$ detections with respect to the true physical ground truth objects $G_i \in \mathcal{G}_\mathcal{E}$, we can introduce object-based precision/recall metrics such that $\mathit{TP} + \mathit{FN} = k$.  This allows us to measure EaRLaP performance when only counting a true positive for each physical ground truth object once and when reconciling detections into object reports across the various pipeline stages.

\subsection{Objectives}
\label{subsec:objectives}
When dealing with harsh unknown environments, it is very unlikely for the robots to be able to fully explore them, especially if there are time constraints. One of Team CoSTAR's strategic goals during SubT was therefore to aim for the highest possible recall so as to avoid the operator missing scores for the true positive objects that the robots encountered.
Aiming for high recall often comes at the cost of poor precision \cite{buckland1994relationship}.
Before CoSTAR adopted the EaRLaP pipeline \cite{terryObjectGasSource2020}, the team (as well as other teams \cite{hudsonHeterogeneousGroundAir2021}) suffered from low precision. During the 2020 Urban Circuit of the DARPA SubT Challenge, the human operator was unable to identify some true positive objects in a timely manner as they were mixed between hundreds of false positives.
Given the many other tasks the operator had to manage (including deploying robots, checking sensor health, setting exploration strategy, etc.), an additional strategic goal was therefore to greatly reduce the number of false positives they had to process for scoring submission in order to save them time for these critical tasks and to maximize precision.
In this section, we aim to demonstrate using the data from the SubT Final Event and the evaluation criteria in Sec. \ref{subsec:eval_criteria}, that EaRLaP was indeed effective at achieving these strategic goals.


\begin{table}[h]
\caption{\label{tab:metrics_detection} Number of TP and FP images/reports at the output of different sub-problems.}
\centering
\scriptsize
\begin{tabular}{lrrrr}
\toprule
 &  
  & \begin{tabular}{@{}c@{}}Detection \\ Output \textsuperscript{1} \end{tabular} 
 &   \begin{tabular}{@{}c@{}}Robot \\ Output \textsuperscript{2} \end{tabular} 
 &  \begin{tabular}{@{}c@{}}Base \\ Output \textsuperscript{3} \end{tabular}  \\ 
 \midrule
\multicolumn{1}{l}{\multirow{3}{*}{\begin{tabular}[c]{@{}l@{}}Preliminary Run\\ (3 robots) \\ (9 true objects) \end{tabular}}} & $\mathit{TP}$ & 4574 & 45 & 10\\
\multicolumn{1}{l}{} & $\mathit{FP}$ & 4490 & 238 & 50\\ \cmidrule{2-5}
\multicolumn{1}{l}{} & time\textsuperscript{4} & 19h & 35min & 7.5min\\ 
\midrule
\multicolumn{1}{l}{\multirow{3}{*}{\begin{tabular}[c]{@{}l@{}}Final Run\\ (3 robots) \\ (4 true objects) \end{tabular}}} & $\mathit{TP}$ & 3699 & 45 & 4\\
\multicolumn{1}{l}{} & $\mathit{FP}$ & 3956 & 260 & 48\\ \cmidrule{2-5}
\multicolumn{1}{l}{} & time\textsuperscript{4} &  16h & 38min & 6.5min\\ 
 \bottomrule
\end{tabular}%

\vspace{1.5mm}
 \scriptsize{See Fig. \ref{fig:prob_decomposition} for the definition of \emph{Detection/Robot/Base Output}}  \\
\vspace{1.5mm}
\scriptsize{\textsuperscript{1} Number of bounding box image detections } \\
\scriptsize{\textsuperscript{2} Reports where many image detections have been reconciled by the robot} \\
\scriptsize{\textsuperscript{3} Groups of reports, reconciled on the base station} \\
\scriptsize{\textsuperscript{4} Estimated time to evaluate each column using an average of 7.5s per image or object 
report} 
\end{table}

\begin{table}[h]
\caption{\label{tab:metrics_object} True Positive ($\mathit{TP}$) and False Positive ($\mathit{FP}$) objects at the output of different sub-problems.}
\centering
\scriptsize
\begin{tabular}{llrrrr}
\toprule
 & 
 & \begin{tabular}{@{}c@{}}Detection \\ Output \end{tabular} 
 &   \begin{tabular}{@{}c@{}}Robot \\  Output \end{tabular} 
 &  \begin{tabular}{@{}c@{}}Base \\  Output \end{tabular} 
 &  \begin{tabular}{@{}c@{}}Operator \\ Output\end{tabular}  \\ \midrule
\multicolumn{1}{l}{\multirow{4}{*}{\begin{tabular}[c]{@{}l@{}}Preliminary Run \\ (3 robots) \\ (9 true objects) \end{tabular}}} & $\mathit{TP}$ & 9 & 9 & 9 & 8 \\
\multicolumn{1}{l}{} & $\mathit{FP}$ & 278 & 99 & 48 & 0 \\ \cmidrule{2-6} 
\multicolumn{1}{l}{} & Recall & 100\% & 100\% & 100\% & 89\% \\
\multicolumn{1}{l}{} & Precision & 3.1\% & 8.3\% & 17.24\% & 100\% \\
\midrule
\multicolumn{1}{l}{\multirow{4}{*}{\begin{tabular}[c]{@{}l@{}}Final Run\\ (3 robots) \\ (4 true objects) \end{tabular}}} & $\mathit{TP}$ & 4 & 4 & 4 & 4 \\
\multicolumn{1}{l}{} & $\mathit{FP}$ & 479 & 113 & 48 & 0 \\ \cmidrule{2-6} 
\multicolumn{1}{l}{} & Recall & 100\% & 100\% & 100\% & 100\% \\
\multicolumn{1}{l}{} & Precision & 0.8\% & 3.4\% & 7.7\% & 100\% \\ 
\bottomrule
\end{tabular}%

\vspace{1.5mm}
\scriptsize{ See Fig. \ref{fig:prob_decomposition} for the definition of \emph{Detection/Robot/Base Output}. Operator Output is the solution submitted by the operator.} 
\end{table}

\subsection{Key Results}
\label{subsec:keyresults}

The competition took place between Sep. 21st--24th, 2021. Over the first three days, competing teams performed in two preliminary runs of 30 minutes each, in which 20 artifacts were hidden. Tables \ref{tab:metrics_detection} and \ref{tab:metrics_object} show results from the second preliminary run. Using EaRLaP, Team CoSTAR scored the most artifacts of all teams during this round, finishing at the top of the score table. CoSTAR proceeded into the final run as one of the favorites to win. The run lasted one hour and consisted of 40 artifacts. Unfortunately, in the final run, our robots encountered challenging conditions and struggled to explore the environment as effectively as in prior runs. Between one Husky and two Spots only four RGB-detectable objects were encountered. The team utilized more robots with diverse sensory capabilities, thus allowing us to score a total of 13 artifacts and achieving a final rank of $5^{th}$ place, but data from these robots could not be retrieved. 

In Tables \ref{tab:metrics_detection} and \ref{tab:metrics_object}, we also show results from some of the most important EaRLaP sub-problems (as illustrated in Fig. \ref{fig:prob_decomposition}) using data from both the preliminary and final runs. Table \ref{tab:metrics_detection} shows results for the image-based metrics discussed in Sec. \ref{subsec:eval_criteria}. In order to produce this table, each image or report from each run was manually labelled to be a true positive ($\mathit{TP}$) or false positive ($\mathit{FP}$).
In the \emph{Detection Output} column we have bounding box detections from our filtered object detector ($f_1$ + $f_2$), while in the \emph{Robot Output} column we have reports where multiple detections have been reconciled together on the robots ($f_4$) and in the \emph{Base Output} column we have groups of reports on the base station ($f_6$). A human operator could theoretically review the data at any of these sub-problems' outputs (although it might not be a good use of their time) therefore Table \ref{tab:metrics_detection}  shows the amount of information that would be exposed to the operator at those stages.  This table however gives no information on the number of actual objects that the operator would encounter, which is what is truly important for the DARPA SubT challenge and the semantic object mapping problem (i.e. the detector can generate 50 detections of the same backpack).

It is thus necessary to introduce Table \ref{tab:metrics_object}, where we counted the number of unique object instances at the output of the different sub-problems and used the object-based metrics discussed in Sec. \ref{subsec:eval_criteria}. To create this table, an annotator had to go through all the images or reports (depending on the sub-problem) and keep track of all unique objects already detected to determine whether to add a new object or not. Additionally, the annotator identified the $\mathit{TP}$ as the objects that we are trying to detect, and all the rest was FP. In addition to being directly related to the goal of the DARPA SubT challenge and semantic mapping problem, using object instances also allows us to break the barrier of different data types at the output of different sub-problems. It thus enables us to compare the performance at various stages of the EaRLaP pipeline, with particular attention being paid to the object-based recall and precision.


From Table \ref{tab:metrics_object}, it can be seen that in both the preliminary and final runs we reached our objective since all $\mathit{TP}$ objects encountered by the robots reached the operator. This was possible due to the wide FOV covered by the cameras and the design of our object detector ($f_1$). During the preliminary run, the operator made an error as they did not recognize a backpack detected at a far distance and in a dark area as a true positive. This may be explained by the fact that this was a practice round and the operator thus was not required to perform at their fullest capacity. This serves as a poignant reminder that although a human's intervention is required to achieve extremely high precision, the operator can also make mistakes. Despite this error, CoSTAR still scored higher than any other team in this particular round. 

As expected, the YOLO CNN found a large number of false positives, shown by the \textit{Detection Output} column in Table \ref{tab:metrics_object}. However, the table shows that as we advance through the sub-functions, the precision increases while the recall is kept at 100\%. Although each sub-problem might only contribute slightly to improving precision, the compounding effect of all sub-problems leads to a 5x increase in precision for the preliminary run and a 9.5x increase for the final run. We believe that achieving such a high precision increase without harming the recall would be quite difficult to achieve with a one-stage filtering function, and we thus recommend a multi-stage pipeline such as ours.

When introducing the seven sub-functions in Sec. \ref{sec:methodology}, we required that some sub-functions reduce the amount of information such that the detected objects could be transmitted on low-bandwidth wireless networks and displayed to the operator. Table \ref{tab:metrics_detection} shows that during the preliminary round the object detector generated more than 9000 detections, but after reconciliation on the robot all of this information was reduced to less than 300 reports, which is low enough to avoid saturating the communication network. After reconciliation on the base station, this is further reduced to just 60 different objects, which is manageable for a human to manually inspect.  

\begin{figure}
    \includegraphics[width=\columnwidth]{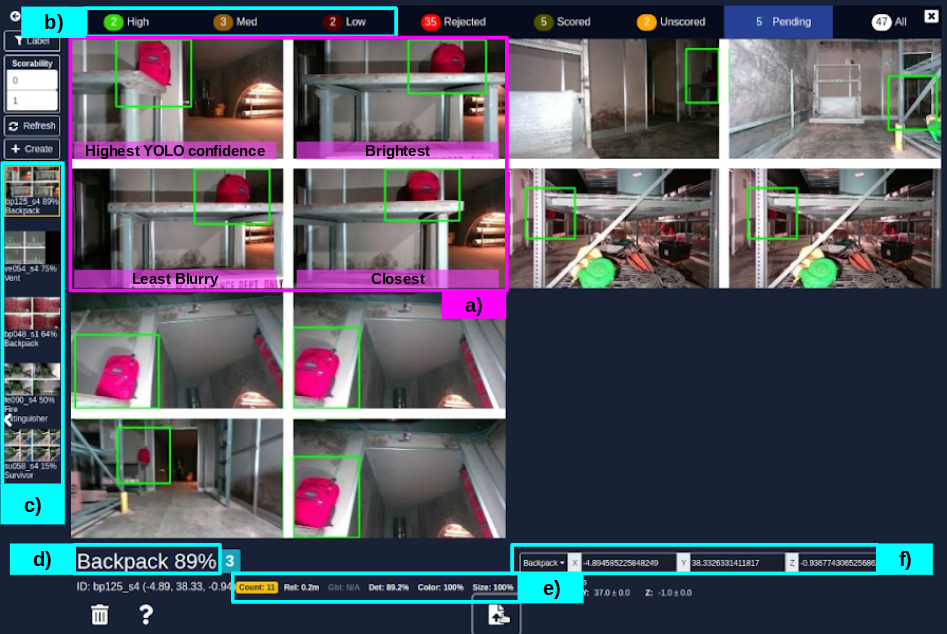}
    \caption{\footnotesize An example of the GUI in the base station shows a candidate submission that contains three artifacts reported from different robots at different times. a) A highlighted report. b) Reports are grouped based on high, median, and low report confidence. c) Reports are ordered by report confidence d) report confidence value and artifact label e) more information on YOLO confidence and color/size scores. f) 3D position of the selected report, where the operator can adjust it before submission. }
    \label{fig:ui}
\end{figure}

We used an efficient GUI, shown in Fig. \ref{fig:ui}, which is optimized such that the operator can review all detections rapidly. We estimate that reviewing and accepting or rejecting an object on the GUI takes between 5 and 10 seconds. Table \ref{tab:metrics_detection} shows that if a human were to sort through all the detections generated by the neural network, it would take almost 19 hours to do so for the preliminary round, which is only 30 minutes long. Despite a significant data reduction after the robot reconciliation, it would still take a human more than 30 minutes to inspect. However, at the end of our pipeline, we managed to reduce the operator's work to 7.5 minutes, which is 25\% of the total run time. Therefore, we successfully achieved our goal. Additionally, thanks to the report scorability ranker ($f_6$), the nine true positive objects are among the first objects displayed to the operator.

\section{CONCLUSION}
\label{sec:conclusions}
This paper studied the task of semantic object mapping using human-in-the-loop multi-robot systems and defined a mathematical formulation of the problem. The proposed EaRLaP pipeline decomposes the problem into sub-problems to achieve high precision and high recall under computational, communication network, and human cognitive load constraints, and ultimately to provide high-quality information for human operators. EaRLaP was implemented by being decomposed into seven sub-functions and deployed by Team CoSTAR in the final events of the DARPA SubT Challenge to perform search-and-rescue in perceptually-degraded environments. The performance showed that EaRLaP maintained high recall under the constraints and pruned false positives effectively to allow the human operator to achieve high precision within time limit.

\bibliographystyle{IEEEtran}
\bibliography{references}

\end{document}